\documentclass{llncs}[11pt]

\usepackage{dcolumn,listings}
\usepackage{endnotes,mathtools,amsmath,amssymb}

\newcommand{\ei}{EI}
\newcommand{\eis}{EIs}

\newcommand{\term}{\ensuremath{\tau}}
\newcommand{\atf}{\ensuremath{\alpha}}

\newcommand{\constr}{\ensuremath{\gamma}}
\newcommand{\constrs}{\ensuremath{\Gamma}}

\newcommand{\events}{\ensuremath{\Xi}}
\newcommand{\sat}{\ensuremath{\mathtt{sat}}}

\newcommand{\conj}{\mbox{\tt{\hspace{1.5mm}\&\hspace{1.5mm}}}}

\newcommand{\aconj}{\mbox{\tt{,}}}
\newcommand{\prolog}{\mbox{\tt{prolog}}}

\newcommand{\rnot}{\mbox{\tt{not}}}

\newcommand{\add}{\mbox{\tt{add}}}
\newcommand{\del}{\mbox{\tt{del}}}

\def\k#1{\mbox{\ensuremath{\mathsf{\mathbf{#1}}} }}
\def\kb#1{\mbox{ \ensuremath{\mathsf{\mathbf{#1}}} }}
\def\kc#1{\mbox{\ensuremath{\mathsf{\mathbf{#1}}}}}
\newcommand{\fulfilledif}{\mbox{ \ensuremath{\mathsf{\mathbf{fulfilled}}}-\ensuremath{\mathsf{\mathbf{if}}} }}
\newcommand{\violatedif}{\mbox{ \ensuremath{\mathsf{\mathbf{violated}}}-\ensuremath{\mathsf{\mathbf{if}}} }}
\newcommand{\sanctiondo}{\mbox{ \ensuremath{\mathsf{\mathbf{sanction}}}-\ensuremath{\mathsf{\mathbf{do}}} }}
\newcommand{\I}{\ensuremath{\mathcal{I}}}

\newcommand{\ctime}{{\tt time}}

\begin{document}


\title{Towards Regulated Deep Learning\\  \vspace{5mm}\small First step in a Roadmap towards Collaborative Knowledge Evolution}

\author{Andr\'es Garc\'ia-Camino}\institute{ORCID:0000-0003-1787-3206}

\maketitle

\begin{abstract}
   Regulation of Multi-Agent Systems (MAS) and Declarative Electronic Institutions (DEIs)   was a multidisciplinary research topic of the past decade involving (Physical and Software) Agents and Law since the beginning, but recently evolved towards News-claimed Robot Lawyer since 2016.  
One of these first proposals of restricting the behaviour of Software Agents was Electronic Institutions.

However, with the recent reformulation of Artificial Neural Networks (ANNs) as Deep Learning (DL), Security, Privacy, Ethical and Legal issues regarding the use of DL has raised concerns in the Artificial Intelligence (AI) Community.
Now that the Regulation of MAS is almost correctly addressed, we propose the Regulation of Artificial Neural Networks as Agent-based Training of a special type of regulated Artificial Neural Network that we call Institutional Neural Network (INN).

The main purpose of this paper is to bring attention to Artificial Teaching (AT) and to give a tentative answer showing a proof-of-concept implementation of Regulated Deep Learning (RDL). This paper introduces the former concept and provides $\mathcal{I}$, a language previously used to model declaratively and extend  Electronic Institutions, as a means to regulate the execution of Artificial Neural Networks and their interactions with Artificial Teachers (ATs).
\end{abstract}


\section{Motivation}

Regulation of Multi-Agent Systems (MAS) and Declarative Electronic Institutions (DEIs) \cite{agc:aamas05}\cite{agc:coin_aamas07}\cite{garcia2010normative}  was a multidisciplinary research topic of the past decade involving (Physical and Software) Agents and Law since the beginning, but recently evolved towards News-claimed Robot Lawyer since 2016.  
One of these first proposals of restricting the behaviour of Software Agents was Electronic Institutions \cite{noriegaphd} \cite{rodriguezphd} \cite{estevaphd}.

However, with the recent reformulation of Artificial Neural Networks (ANNs) as Deep Learning (DL) \cite{Lecun2015}, Security, Privacy, Ethical and Legal issues regarding the use of DL has raised concerns in the Artificial Intelligence (AI) Community \cite{Dignum2018}.

Now that the Regulation of MAS is almost correctly addressed, we propose the Regulation of Artificial Neural Networks as Agent-based Training of a special type of regulated Artificial Neural Network that we call Institutional Neural Network (INN).

The main purpose of this paper is to bring attention to Artificial Teaching (AT) and to give a tentative answer showing a proof-of-concept implementation of Regulated Deep Learning (RDL). 

This paper introduces the former concept and provides $\mathcal{I}$ \cite{agc:coin_aamas07}, a language previously used to model and extend  Electronic Institutions\cite{estevaphd}, as a means to regulate the execution of Artificial Neural Networks and their interactions with Artificial Teachers (ATs).

The structure of this position article is as follows, in section \ref{proposal} the need of Agent-based Training for DL (ABT4DL) is introduced. In Section \ref{introducingI} the fore-mentioned language is introduced and, Section \ref{defs} presents some preliminary definitions for this work. The normative language that includes the tentative answer is presented in section \ref{language}. Section \ref{semantics} presents the Semantics of the language and section \ref{opsemantics} its Operational Semantics. Some examples of Artificial Neural Network modeling are presented in section \ref{example-soup}.
 And finally, sections \ref{conclusions} and \ref{future-work} presents some conclusions and possible future work to continue this research, respectively.

\section{Agent-based Training for Deep Learning}\label{proposal}

Current ANNs, and subsequently Deep Learning, proposals are Data-driven: they model an a priori unknown function by processing large amounts of data \cite{Lecun2015} without automatically controlling the source of this data, neither how one could certify the proper use of the data. 

We introduce Agent-based Training for Deep Learning (ABT4DL) as a new paradigm for Automated training of Artificial Neural Networks. However, we envisage that such training should be regulated and mediated, thus, as we propose agents as artificial data providers with an intuitive role of Artificial Teacher (AT), using a Regulated MAS as follows in order to regulate and mediate the behaviour of these Artificial Teachers interacting with Artificial Neural Networks residing in the state-of-affairs of a Declarative Electronic Institution \cite{garcia2010normative}.

\section{Introducing a Regulation Language}\label{introducingI}

The proposed language in this work is based in \emph{Event-Condition-Action} (ECA) rules, that perform the specified institutional actions when the given agent events occur and when the prespecified conditions hold. The performance of actions add or remove atomic formulae thus triggering another type of rule: the \emph{if-rules}, that are standard production rules as usually found in Expert Sytems \cite{vianu97}. Above these two types of rules a new types of rules are placed: \emph{ignore-rules}, that ignore a set of simultaneous events, \emph{force-rules}, that generate a set of events on the occurrence of a given set of events satisfying certain conditions, and \emph{prevent-rules}, that ignore the execution of \emph{ECA} or \emph{if-then} rules if certain formulae hold in the current state and another given set of formulae hold in the calculation of the next state.

\subsection{Preliminary Definitions}\label{defs}
Some basic concepts are now introduced.  The building blocks of the
language are \emph{terms}:
\begin{definition}
\label{def:term}
A term, denoted as $\term$, is
\begin{itemize}
\item Any variable $x,y,z$ (with or without subscripts) or
\item Any construct $f^n(\term_1,\ldots,\term_n)$,
      where $f^n$ is an n-ary function symbol and $\term_1,
      \ldots, \term_n$ are terms.
\end{itemize}
\end{definition}
 Terms $f^0$
stand for \emph{constants} and will be denoted as $a,b,c$ (with or
without subscripts).  Numbers and arithmetic
functions to build terms are used ; arithmetic functions may appear infix,
following their usual conventions. Prolog's convention
\cite{apt97} is used using strings starting with a capital letter to represent
variables and strings starting with a small letter to represent
constants. Some examples of terms are $\mathit{Price}$ (a variable)
and $\mathit{send}(a,B,Price \times 1.2)$ (a function).  The definition of \emph{atomic formulae} is as follows:
\begin{definition}
\label{def:atf}
An atomic formula, denoted as \atf, is any construct
$p^{n}(\term_1, \allowbreak \ldots, \allowbreak
\term_{n})$, where $p^{n}$ is an $n$-ary predicate
symbol and $\term_1, \allowbreak \ldots, \allowbreak
\term_{n}$ are terms.
\end{definition}
When the context makes it clear what $n$ is, it is dropped. $p^0$
stands for propositions. Arithmetic relations
(\emph{e.g.}, $=$, $\neq$, and so on) are employed as predicate symbols, and these
will appear in their usual infix notation. Atomic
formulae built with arithmetic relations to represent
\emph{constraints} on variables are used -- these atomic formulae have a
special status, as it is explained below.  The definition of 
constraints in this work, a subset of atomic formulae is:
\begin{definition}
\label{def:constraints} A constraint $\constr$ is a binary atomic formula
 $\tau \vartriangleleft \tau ,$ where
$\vartriangleleft\, \in\{=,\neq,>,\geq,<,\leq\}$.
\end{definition}
$\Gamma = \{\gamma_1,\ldots,\gamma_n\}$ is used as a set of constraints.
A state of affairs is a set of atomic formulae, representing (as shown
below) the normative positions of agents, observable agent attributes
and the state of the environment\footnote{It is referred to the \emph{state
of the environment} as the subset of atomic formulae representing
observable aspects of the environment in a given point in time.}.
\begin{definition}
\label{def:state-of-affairs}
A state of affairs $\Delta = \{\atf_0:\constrs_0, \ldots, \atf_n:\constrs_n\}$ is a a
finite and possibly empty set of implicitly, universally quantified
atomic formulae $\atf_i$ restricted by a possibly empty set of constraints $\constrs_i$,  $0\leq i\leq n$.
When the set of constraints is empty, it is written just $\atf_i$.
\end{definition}

\section{\I: A Language for Institutional Neural Networks}
\label{language} 

 This section introduces a
rule language for the regulation and management of 
concurrent events generated by a
population of agents. 

\begin{figure*}[htb]
\begin{center}
\begin{large}
$$
\fbox{$\Delta_0$}
\Rrightarrow
\begin{array}{|c|}
  \hline
\hspace{35mm}
  \;\vspace*{-2mm} \\
  \hspace{-7mm}\fbox{$\Delta_0$}\hspace*{-8mm} \\
  \;\vspace*{-2mm} \\
  \hspace{-12mm} \events^0_1,\cdots,\events^0_n \hspace*{-12mm} \\ \hline
  \multicolumn{1}{c}{
    \begin{array}{ccc}
      \hspace{-1mm}\updownarrow & & \hspace{0mm}\updownarrow \\
      \hspace{-1mm}\mathit{ag}_1 & \hspace{0.2mm}\cdots \hspace{2mm}&
                                   \hspace{2mm}\mathit{ag}_n\hspace*{-2mm}
    \end{array}}
\end{array}
\stackrel{*}{\rightsquigarrow} \fbox{$\Delta_1$}
\Rrightarrow
\begin{array}{|c|} \hline
\hspace{35mm}
  \;\vspace*{-2mm} \\
  \hspace{-7mm}\fbox{$\Delta_1$}\hspace*{-8mm} \\
  \;\vspace*{-2mm} \\
  \hspace{-11mm} \events^1_1,\cdots,\events^1_m \hspace*{-12mm} \\ \hline
  \multicolumn{1}{c}{
    \begin{array}{ccc}
      \hspace{-1mm}\updownarrow & & \hspace{0mm}\updownarrow \\
      \hspace{-1mm}\mathit{ag}_1 & \hspace{0.2mm}\cdots \hspace{2mm}&
                                   \hspace{2mm}\mathit{ag}_m\hspace*{-2mm}
    \end{array}}
\end{array}
\stackrel{*}{\rightsquigarrow}
\cdots
$$
\end{large}
\caption{Semantics as a Sequence of $\Delta$'s}
\label{fig:updates2}
\end{center}
\end{figure*}

Figure~\ref{fig:updates2} depicts the computational model used: n initial state of affairs $\Delta_0$ (possibly empty) is
offered (represented by ``$\Rrightarrow$'') to a set of agents
($ag_1,\cdots,ag_n$). These agents can add their speech acts ($\events^0_1,\cdots,\events^0_n$) to
the state of affairs (via ``$\updownarrow$'').
 $\events^t_i$ is the (possibly empty) set of speech acts added by agent $i$ at state of affairs
 $\Delta_t$. 
 After an established amount of time, an exhaustive application of rules is performed(denoted by
``$\stackrel{*}{\rightsquigarrow}$'') to the modified state, yielding
a new state of affairs $\Delta_1$. This new state will, in its turn,
be offered to the agents for them to add their utterances, and the same
process will go on.

One goal of the $\I$ language is to specify the effects of concurrent events and this is achieved with Event-Condition-Action (ECA) rules.
Intuitively, an ECA-rule means that whenever  the events occur and the conditions hold then the actions are applied. These actions consist of the addition and removal of atomic formulae from the state of affairs.
ECA-rules are checked in parallel and they are executed only once without chaining.

If-rules are similar to rules in standard production systems, if the conditions hold then the actions are applied. They are implemented with a forward chaining mechanism: they are executed sequentially until no new formula is added or removed. 

Ignore-rules are used for ignoring events when the conditions hold in order to avoid unwanted behaviour.
Similarly, prevent-rules are used for preventing some conditions to hold in the given situations. In order to prevent unwanted states, events causing such unwanted states are ignored.
Force-rules generate events and execute actions as consequence of other events and conditions. 

Sanctions over unwanted events can be carried out with ECA-rules. For instance, they can decrease the credit of one agent by 10 if she generates a certain event.

\begin{figure*}[htb]
\center

\fbox{
$
\begin{array}{rcl}
\mbox{$ECA$-$Rule$} & ::= & \k{on} set\_of\_events \kb{if} conditions \kb{do} actions
\\
\mbox{$if$-$Rule$} & ::= & \k{if} conditions \kb{do} actions
\\
\mbox{$ignore$-$Rule$} & ::= & \k{ignore} set\_of\_events \kb{if} conditions 
\\
\mbox{$prevent$-$Rule$} & ::= & \k{prevent} conditions \kb{if} conditions 
\\
\mbox{$force$-$Rule$} & ::= & \k{force} set\_of\_events \kb{on} set\_of\_events\\
&& \k{if} conditions \kb{do} actions
\\
set\_of\_events & ::= & events \,|\, \emptyset
\\
events & ::= & atomic\_formula, events \,|\,
           atomic\_formula \\
conditions & ::= & conditions\conj conditions \,|\,
                     \rnot(conditions) \\
                     &\,|\, & \sat(set\_of\_constraints)      \,|\, constr\_formula	\\	
                     &\,|\, & \mathtt{seteq}(set\_of\_constraints,set\_of\_constraints)\\		 
                     &\,|\,& constraint \in set\_of\_constraints\\
                     &\,|\,& \ctime(number) \,|\, \mathtt{true} 
\\
constr\_formula & ::= & atomic\_formulae\\
			& \,|\,& atomic\_formulae : set\_of\_constraints
\\
\mathit{actions} & ::= & action\aconj actions \,|\, action
\\
action & ::= & \add (constr\_formula) \,|\, \del( constr\_formula)
\end{array}
$ 
} \caption{Grammar for $\I$}\label{I:fig:grammar} 
\end{figure*}

Figure \ref{I:fig:grammar} shows the grammar for \I, \emph{i.e.} the syntax of the five type of rules proposed: ECA-rules, if-rules, ignore-rules, prevent-rules and force-rules. 

 ECA rules specify the effect of a set of events, \emph{i.e.} a set of atomic formulae, if the conditions hold. This effect is specified by means of a sequence of actions namely addition and removal of constrained formulae. A constrained formulae is an atomic formula that may be followed by a set of arithmetical constraints using the syntax presented in Def. \ref{def:constraints}. Furthermore, by $conditions$ is meant  one or more possibly negated conditions. Then, a condition may be a constrained formula, the  $\sat$ predicate that checks that a set of constraints is satisfiable,  the {\tt seteq} predicate that checks if two sets are equal, the $\ctime$ predicate that checks current time or the {\tt true} constant that always hold.

If-rules specify the logical consequence if the conditions hold by means of a sequence of actions.
Ignore-rules specify the set of events that should be ignored if the conditions hold.
Similarly, prevent-rules specify the conditions that should not hold if some conditions hold.
Finally, force-rules specify a set of new events that are generated on the occurrence of a set of events and the satisfaction of a sequence of conditions.
Furthermore, it also specifies a sequence of actions to perform if the rule is triggered.

An extra kind of rule, called expectation-rules, and shown in Fig. \ref{fig:expectation-rules}, that generate and remove expectations of events might be added. If the expectation fails to be fulfilled then some sanctioning or corrective actions are performed.

\begin{figure*}[htb]
\begin{eqnarray*}
\centering
\mbox{$expectation$-$Rule$} & ::= & \k{expected} event \kb{on} set\_of\_events \kb{if} conditions\\
&& \!\!\fulfilledif conditions' \violatedif conditions''\\
&& \!\!\sanctiondo  actions\\
\end{eqnarray*}

    \caption{Expectation Rule}
    \label{fig:expectation-rules}
\end{figure*} 

However, each expectation rule are equivalent to the following rules:

\begin{figure*}
    \centering
{
\begin{eqnarray}
\k{on} set\_of\_events \kb{if} conditions \kb{do} \add(exp(event))\label{eq:add-exp}\\
\k{if}  exp(event) \wedge conditions' \kb{do} \del(exp(event))\label{eq:del-exp}\\
\k{if} exp(event) \wedge conditions'' \kb{do} \del(exp(event))\aconj actions \label{eq:exp-sanction}
\end{eqnarray}
}

    \caption{Expectation rule semantics expressed with 3 rules}
    \label{fig:exp-ex}
\end{figure*}

In Fig. \ref{fig:exp-ex}, Rule \ref{eq:add-exp} and \ref{eq:del-exp} respectively adds and removes an expectation whenever the events have occurred and the conditions hold.
Rule \ref{eq:exp-sanction} cancels the unfulfilled expectation and sanctions an agent for the unfulfilled expectation by executing the given $actions$ whenever some $conditions$ hold.

\section{Semantics}\label{semantics}

Instead of basing the \I\ language on the standard deontic notions, two types of prohibitions and two types of obligations are included. In the language, ECA-rules determine what is possible to perform, i.e. they establish the effects (including sanctions) in the institution after performing certain (possibly concurrent) events. ECA-rules can be seen as conditional count-as rules: the given events count as the execution of the actions in the ECA-rule if the conditions hold and the event is not explicitly prohibited. 
As for the notion of permission, all the events are permitted if not explicitly prohibited.
The notion of an event being prohibited may be expressed depending on whether that event has to be ignored or not. If not otherwise expressed, events are not ignored.
Likewise, the notion of a state being prohibited may be specified depending on whether that state has to be prevented or not. By default, states are not prevented.
Obligations are differentiated in two types: expectations, which an agent may not fulfill, and forced (or obligatory) events, which the system takes as institutional events even they are not actually performed by the agents.

Each set of ECA-rules generates a labelled transition system
$\langle \mathcal{S},\mathcal{E},\mathcal{R} \rangle$ where $\mathcal{S}$ is  a set of states, each state in $\mathcal{S}$ is a set of atomic formulae,
$\mathcal{E}$ is a set of events,  and $\mathcal{R}$ is a $\mathcal{S}\times 2^\mathcal{E} \times \mathcal{S}$ relationship indicating that
whenever a set of events occur in the former state, then there is a transition to the subsequent state.

Ignore-rules avoid executing any transition that contains in its labelling the events that appear in any ignore-rule. For instance, having a rule $\k{ignore} \alpha_1 \allowbreak \kb{if} true$ would avoid  executing the transitions labelled as $\{\alpha_1\}$, $\{\alpha_1, \alpha_2\}$ and $\{\alpha_1, \alpha_2, \alpha_3\}$. However, having a rule $\k{ignore} \alpha_1, \alpha_2 \kb{if} true$ would avoid executing $\{\alpha_1, \alpha_2\}$ and $\{\alpha_1, \alpha_2,\alpha_3\}$ but not $\{\alpha_1\}$.

 Prevent-rules ignore all the actions in an ECA-rule if it brings the given formulae about. For example, suppose that we have  \[\k{prevent} q_1 \kb{if} true\] 
 along with ECA-rules \ref{eq:add-p}, \ref{eq:add-q} and \ref{eq:add-r} below. After the occurrence of events $\alpha_1$ and $\alpha_2$ and  since $q_1$ is an effect of event $\alpha_2$, all the actions in ECA-rule \ref{eq:add-q} would be ignored obtaining a new state where $p$ and $r$ hold but neither $q_1$ nor $q_2$.
 
\begin{eqnarray}
\k{on} \alpha_1 \kb{if} true \kb{do} \add(p)\label{eq:add-p}\\ 
\k{on} \alpha_2 \kb{if} true \kb{do} \add(q_1)\aconj\add(q_2) \label{eq:add-q}\\ 
\k{on} \alpha_1, \alpha_2 \kb{if} true \kb{do} \add(r) \label{eq:add-r} 
\end{eqnarray}

Force-rules generate events during the execution of the transition system. However, the effects of such events are still specified by ECA-rules and subject to prevent and ignore-rules.

\section{Operational Semantics}\label{opsemantics}

As shown in figure \ref{fig:updates2}, the semantics of 
rules are presented as a relationship between states of affairs: rules map an
existing state of affairs to a new state of affairs.  This section defines this relationship. 
The definitions below rely on the concept of {\em substitution},
that is, the set of values for variables in a computation
\cite{apt97,fitting90}:
\begin{definition}
A substitution $\sigma=\{x_0/\term_0, \ldots,
x_n/\term_n\}$ is a finite and possibly empty set of pairs
$x_i/\term_i$, $0 \leq i \leq n$.
\label{def:substitution}
\end{definition}
\begin{definition}
The application of a substitution to an atomic formulae $\atf$ possibly restricted by a set of constraints $
\{\constr_0,\ldots,\constr_m\}$ is as
follows:
\begin{enumerate}
\item $c\cdot\sigma = c$ for a constant $c$;\;
\item $x\cdot\sigma = \term\cdot\sigma$ if
$x/\term\in\sigma$; otherwise $x\cdot\sigma = x$;\;
\item $p^n(\term_0,\ldots,\term_n)\cdot\sigma =
       p^n(\term_0\cdot\sigma,\ldots,\term_n\cdot\sigma)$;
\item $p^n(\term_0,\ldots,\term_n):\{\constr_0,\ldots,\constr_m\}\cdot\sigma =
       p^n(\term_0\cdot\sigma,\ldots,\term_n\cdot\sigma):\{\constr_0\cdot\sigma,\ldots,\constr_m\cdot\sigma\}$.
\end{enumerate}
\end{definition}
\begin{definition}
The application of a substitution to a sequence is the sequence of the application of the
substitution to each element:
$\langle\atf_1,\ldots,\atf_n\rangle\cdot\sigma = \langle\atf_1\cdot\sigma,\ldots,\atf_n\cdot\sigma\rangle$
\label{def:substitution-appl}
\end{definition}

The semantics of the conditions are now defined, that
is, when a condition holds:
\begin{definition}
\label{def:s-l2}
Relation $\mathbf{s}_l(\Delta,C,\sigma)$ holds between state
$\Delta$, a condition $C$ in an \kc{if} clause and a 
substitution $\sigma$ depending on the format of the condition:
\begin{enumerate}

\item $\mathbf{s}_l(\Delta,C\,\conj\,C',\sigma)$
      holds iff $\mathbf{s}_l(\Delta, C, \sigma')$ and
      $\mathbf{s}_l(\Delta, \allowbreak C'\cdot\sigma', \sigma'')$ hold and
      $\sigma=\sigma'\cup\sigma''$.

\item $\mathbf{s}_l(\Delta,\rnot(C),\sigma)$ holds iff
      $\mathbf{s}_l(\Delta, C ,\sigma)$ does not hold.

\item $\mathbf{s}_l(\Delta,{\tt seteq}(L,L2),\sigma)$ holds iff $L\subseteq L2$, $L2\subseteq L$ and $|L|=|L2|$.

\item $\mathbf{s}_l(\Delta,\sat(constraints),\sigma)$ holds iff\\ $\mathit{satisfiable}(constraints\cdot\sigma)$ hold.

\item $\mathbf{s}_l(\Delta,\constr \in \constrs,\sigma)$ holds iff $(\constr\cdot\sigma) \in (\constrs\cdot\sigma)$.

\item $\mathbf{s}_l(\Delta,\mathtt{time}(T),\sigma)$ holds iff  current time is $T$.

\item $\mathbf{s}_l(\Delta,{\tt true},\sigma)$ always holds.

\item $\mathbf{s}_l(\Delta,constr\_formula,\sigma)$ holds iff\\
      $constr\_formula\cdot\sigma\in \Delta$.

\end{enumerate}
\end{definition}

Case 1 depicts the semantics of atomic formulae and how their
individual substitutions are combined to provide the semantics for a
conjunction. Case 2 introduces negation by failure.
Case 3 compares if two lists have the same elements possibly in different order.
Case 4 checks if a set of constraints is satisfiable.
Case 5 checks if a constraint belongs to a set of constraints.
Case 6 checks if $T$ is current time.
Case 7 gives semantics to the keyword {\tt true}.
Case 8 holds when an possibly constrained, atomic formulae $constr\_formula$ is part of the state
of affairs. 

The semantics of the actions of a rule are now defined:
\begin{definition}
\label{def:s-r}
Relation $\mathbf{s}_r(\Delta,A,\Delta')$ mapping a
state $\Delta$, the action section of a rule and a new state $\Delta'$ is defined as:
\begin{enumerate}
\item
      $\mathbf{s}_r(\Delta,(A\aconj As),\Delta')$
      holds iff both $\mathbf{s}_r(\Delta,A,\allowbreak
      \Delta_1)$ and $\mathbf{s}_r(\Delta_1, \allowbreak As,
      \allowbreak \Delta')$ hold.

\item $\mathbf{s}_r(\Delta,\add(constr\_formula),\Delta')$ holds iff 
\begin{enumerate}
	\item $constr\_formula \not\in\Delta$ and $\Delta'=\Delta \cup \{constr\_formula\}$ or;
	\item $\Delta'=\Delta$.
\end{enumerate}
\item $\mathbf{s}_r(\Delta,\del(constr\_formula),\Delta')$ holds iff
\begin{enumerate}
	\item      $constr\_formula \in\Delta$ and $\Delta'=\Delta\setminus\{constr\_formula\}$ or;
	\item $\Delta'=\Delta$.
\end{enumerate}	
\end{enumerate}
\end{definition}
Case 1 decomposes a conjunction and builds the new state by merging
the partial states of each update. Case 2 and 3 cater respectively for the insertion and removal of
atomic formulae $\alpha$.

Relation $check_{prv}$  checks if there is no prevent-rule that has been violated, i.e., it is not the case that all the conditions of any prevent-rule hold in the state of affairs $\Delta'$. It checks whether $\Delta'$  contain all the conditions of each prevent-rule or not, if $\Delta$ also contain the given conditions.

\begin{definition}
\label{def:check_prv}
Relation $check_{prv}(\Delta,\Delta',PrvRules)$ mapping 
$\Delta$, the state before applying updates, $\Delta'$, the state after applying updates, and a sequence $PrvRules$ of prevent-rules, holds iff an empty set is the largest set of conditions $C$ such that:
 \[\begin{array}{lll} p= \k{prevent} C \kb{if} C', p \in PrvRules,\\ and\ \mathbf{s}_l(\Delta,C'), \mathbf{s}_l(\Delta',C)\ hold\end{array}\]
\end{definition}

\begin{definition}
\label{def:fire}
$fire(\Delta,\ PrvRules,\ \k{if} C \kb{do} A,\ \Delta')$, relation mapping a
state $\Delta$, a sequence $PrvRules$ of prevent-rules, an if-rule and a new state $\Delta'$ holds iff $fired(C,A)$ starts to hold, $\mathbf{s}_r(\Delta,A,\Delta')$ and\\ $check_{prv}(\Delta,\Delta',PrvRules)$ hold.
\end{definition}

Relation $can\_fire$ checks whether the conditions of a given if-rule hold and the rule after applying substitution $\sigma$ has not been already fired.

\begin{definition}
\label{def:can_fire}
Relation $can\_fire(\Delta,\ \k{if} C \kb{do} A,\ \sigma)$ mapping a
state $\Delta$ an if-rule and a substitution $\sigma$ holds iff $\mathbf{s}_l(\Delta,C,\sigma)$ holds and $fired(C\cdot\sigma, A\cdot\sigma)$ does not hold.
\end{definition}

Relation $resolve$ determines the rule that will be fired by selecting the first rule in the list.

\begin{definition}
\label{def:resolve}
$resolve(RuleList, SelectedRuleList)$, relation mapping a
list of if-rules and a selected if-rule list holds iff 
\begin{enumerate}
\item $RuleList=\langle\rangle$ and $SelectedRuleList=\langle\rangle$; or
\item $RuleList=\langle r_1,\cdots,r_n \rangle$ and $SelectedRuleList=\langle r_1 \rangle$.
\end{enumerate}
\end{definition}

Relation $select\_rule$ determines the rule that will be fired by selecting all the rules that can fire and resolving the conflict with relation $resolve$.

\begin{definition}
\label{def:select_rule}
Relation $select\_rule(\Delta,IfRulesList,$\\$SelectedRuleList)$ mapping a state of affairs $\Delta$, a list of if-rules and a selected if-rule list holds iff $Rs$ is the largest set of rules $R\in Rs$, $Rs\subseteq IfRulesList$ such that $can\_fire(\Delta,R,\sigma)$; 
$resolve(Rs,SR)$ hold and $SelectedRuleList= SR\cdot\sigma$.
\end{definition}

Relation $\mathbf{s}_{if}$ determines the new state of affairs after applying a set of if-rules to a initial state of affairs taking into account a set of prevent-rules.

\begin{definition}
\label{def:s_if}
$\mathbf{s}_{if}(\Delta,IfRules,PrvRules,\Delta')$, relation mapping a state of affairs $\Delta$, a list of if-rules, a list of prevent-rules and a new state of affairs holds iff
\begin{enumerate}
\item $select\_rule(\Delta,IfRules,R)$ hold, $R\neq\langle\rangle$,\\ $fire(\Delta,PrvRules,R,\Delta'')$ and\\ $\mathbf{s}_{if}(\Delta'',\allowbreak IfRules,PrvRules,\Delta')$ hold; or
\item $select\_rule(\Delta,IfRules,R)$ hold, $R=\langle\rangle$; or
\item $\mathbf{s}_{if}(\Delta,IfRules,PrvRules,\Delta')$ hold.
\end{enumerate}
\end{definition}

Relation $ignored$ determines that a set of events which occurred have to be ignored taking into account a list of ignore-rules.

\begin{definition}
\label{def:ignored}
Relation $ignored(\Delta,\Xi,E,IgnRules)$ mapping a state of affairs $\Delta$, a list $\Xi$ of events that occurred, a list of events in a ECA-rule and a list of ignore-rules holds iff
$i=\k{ignore} E' \kb{if} C$, $i \in IgnRules$, $E'\subseteq \Xi$, $E\cap E' \neq \emptyset$ and $\mathbf{s}_l(\Delta,C)$ holds.
\end{definition}

Relation $\mathbf{s}'_r$ uses $\mathbf{s}_r$ first and then $\mathbf{s}_{if}$ in order to activate the forward chaining.
 
\begin{definition}
\label{def:s_prime_r}
Relation $\mathbf{s}'_r(\Delta,IfRules,PrvRules,$\\$ActionList,\Delta')$ mapping a state of affairs $\Delta$, a list of if-rules, a list of prevent-rules, a list of actions and a new state of affairs holds iff any of the conditions below hold:
\begin{enumerate}
\item $ActionList=\langle\rangle$ and $\Delta'=\Delta$; or
\item $ActionList=\langle a_1,\cdots,a_n\rangle$, $\mathbf{s}_r(\Delta,a_1,\Delta'')$,\\ $check_{prv}(\Delta,\Delta'',PrvRules)$, \\
$\mathbf{s}_{if}(\allowbreak\Delta'',\allowbreak IfRules,PrvRules,\Delta''')$ and\\
$\mathbf{s}'_r(\Delta''',IfRules,PrvRules,\allowbreak \langle a_2,\cdots \allowbreak ,a_n\rangle,\Delta')$\\ hold; or
\item $\mathbf{s}'_r(\Delta,IfRules,PrvRules,\langle a_2, \cdots, a_n\rangle, \Delta')$.
\end{enumerate}
\end{definition}

Relation $\mathbf{s}_{eca}$ calculates the new state of affairs $\Delta'$ from an initial state $\Delta$ and a set $\Xi$ of events that occurred applying a list of ECA-rules, if-rules, ignore-rules and prevent-rules.

\begin{definition}
\label{def:s_eca}
Relation $\mathbf{s}_{eca}(\Delta,\Xi,ECARules,$\\$IfRules,IgnRules,PrvRules,\allowbreak \Delta')$ mapping a state of affairs $\Delta$, a list $\Xi$ of events that occurred, a list of ECA-rules, a list of if-rules, a list of ignore-rules, a list of prevent-rules, and a new state of affairs holds iff:
\begin{itemize}
\item  $As$ is the largest set of actions $A'=A\cdot\sigma$ in an ECA-rule $r=\k{on} E \kb{if} C \allowbreak \kb{do} A$ such that:
\begin{itemize} 
\item $r\in ECARules$, $E\cdot\sigma'\subseteq\Xi$, $\mathbf{s}_l(\Delta,C,\sigma'')$ hold, 
\item $ignored(\Delta,\Xi,E,IgnRules)$ does not hold and
\item $\sigma=\sigma'\cup\sigma''$; and 
\end{itemize}
\item $\mathbf{s}'_r(\Delta,IfRules,PrvRules,As,\Delta')$ hold.
\end{itemize}
\end{definition}

Relation $\mathbf{s}_{force}$ calculates the new state of affairs $\Delta'$ and the new set $\Xi'$ of occurred events from an initial state $\Delta$ and a set $\Xi$ of events that occurred applying a list of if-rules, ignore-rules, prevent-rules and force-rules.

\begin{definition}
\label{def:s_force}
Relation $\mathbf{s}_{force}(\Delta,\Xi,IfRules,$\\$IgnRules,PrvRules,\allowbreak FrcRules,\Xi',\Delta')$ mapping a state of affairs $\Delta$, a list $\Xi$ of events that occurred, a list of if-rules, a list of ignore-rules, a list of prevent-rules, a list of force-rules, a new list of events that occured and a new state of affairs holds iff:
\begin{itemize}
\item $EAs$ is the largest set of tuples $\langle FE\cdot\sigma,A\cdot\sigma \rangle$ of forced events and actions in a force rule $fr=\k{force} FE \kb{on} E \kb{if} C \allowbreak \kb{do} A$ such that 
\begin{itemize}
\item $fr\in FrcRules$, $E\cdot\sigma'\subseteq \Xi$, $\mathbf{s}_l(\Delta,C,\sigma'')$ holds,
\item $ignored(\Delta,\Xi,E, \allowbreak IgnRules)$ does not hold and 
\item $\sigma=\sigma'\cup\sigma''$;
\end{itemize}
\item $Es$ is the largest set of forced events $Ev$ such that $\langle Ev,A\rangle\in EAs$;
\item $\Xi'=\Xi \cup Es$;
\item $As$ is the largest set of actions $A$ such that $\langle Ev,A\rangle\in EAs$; and
\item $\mathbf{s}'_r(\Delta,IfRules,PrvRules,As,\Delta')$ holds.
\end{itemize}
\end{definition}

Relation $\mathbf{s}^*$ calculates the new state of affairs $\Delta'$ from an initial state $\Delta$ and a set $\Xi$ of events that occurred applying a list of ECA-rules, if-rules, ignore-rules, prevent-rules and force-rules.

\begin{definition}
\label{def:s^*}
Relation $$\mathbf{s}^{*}(  \Delta,  \Xi,  ECARls,   IfRls,  IgnRls,  PrvRls,  FrcRls,  \Delta')$$
mapping a state of affairs $\Delta$, a list $\Xi$ of events that occurred, a list of ECA-rules, a list of if-rules, a list of ignore-rules, a list of prevent-rules, a list of force-rules and a new state of affairs holds iff:
\begin{itemize}
\item $Cs$ is the largest set of conditions $C$ such that $fired(C,A)$ stop holding;
\item $fired(false,false)$ starts to hold,
\item $\mathbf{s}_{if}(\Delta,IfRls,PrvRls,\Delta'')$,
\item $\mathbf{s}_{force}(\Delta'',\Xi,IfRls,IgnRls,PrvRls, FrcRls,\Xi',\\\Delta''')$ and
\item $\mathbf{s}_{eca}(\Delta''',\Xi', ECARls,\allowbreak IfRls, IgnRls, PrvRls,\Delta')$\\ hold.
\end{itemize}
\end{definition}

\section{Modelling Neural Networks with rules}
 \label{example-soup}
  
  In this section we introduce how to use \I\ rule-language for modelling Artificial Neural Networks. We start with the more simpler unit of processing in Artificial Neural Networks, \emph{i.e.} Perceptrons.
  
  We will rely on two special type of events, \emph{inputs} specified as $i(x_i,l_j)$, with the intuitive meaning that an agent provided $x_i$ as the input for layer $l_j$; and \emph{outputs} specified as $o(x_i,l_j)$ with the intuitive meaning that $x_i$ is an output for $l_j$.

\subsection{Modelling Multi-layer Perceptrons as rules}  
  
We begin this section presenting an example on how to use the $\I$ language in order to specify a Perceptron:
 
 \begin{figure*}
     \centering
{ 
 \begin{equation}
 \begin{array}{lll}
 &\k{on}& i(X_1,1),\cdots\, i(X_n, 1)\\
  &\kb{if}& \prolog(calculate(X_1,\cdots,X_n,Y)) \\
  &\kb{do}& \add(o(Y,1)) \label{eq:perceptron}
\end{array}
\end{equation}
}

     \caption{Example of one layer}
     \label{fig:one-layer}
 \end{figure*}

In Fig. \ref{fig:one-layer}, Rule \ref{eq:perceptron} translates a set of $n$ inputs, provided by Agents, to one output by means of a calculation implemented in prolog by a $calculate$ predicate.

For specifying multiple layers of Perceptrons, we envisage several rules (Perceptrons) forward chaining in several activations:

 \begin{figure*}
     \centering
{ 
 \begin{eqnarray}
 &\begin{array}{lll}
 &\k{on}& i(X_1,1),\cdots\, i(X_n, 1)\\
  &\kb{if}& \prolog(calculate(X_1,\cdots,X_n,Y_1)) \\
  &\kb{do}& \add(o(Y_1,1)) \label{eq:perceptron11}
\end{array}&\\
\cdots&&\\
& \begin{array}{lll}
 &\k{on}& i(X_j,1),\cdots\, i(X_k, 1)\\
  &\kb{if}& \prolog(calculate(X_j,\cdots,X_k,Y_i)) \\
  &\kb{do}& \add(o(Y_i,1)) \label{eq:perceptron12}
\end{array}&\\
& \begin{array}{lll}
 &\k{if}& o(X_1,1)\conj \cdots \conj o(X_i, 1)\\
  &\conj& \prolog(calculate(X_1,\cdots,X_i,Y_{i+1})) \\
  &\kb{do}& \add(o(Y_{i+1},2)) \label{eq:perceptron2}
\end{array}&
\end{eqnarray}
}
     \caption{Example of second layer}
     \label{fig:second-layer}
 \end{figure*}

For space restrictions, we model a basic two-layer Multi-layer Perceptron (MLP) where $i$ Perceptrons are defined by rules from \ref{eq:perceptron11}--\ref{eq:perceptron12} in Fig. \ref{fig:second-layer}. The last rule gathers the outputs from the previous layer and calculates the final output (of layer 2).

This basic example illustrates expressiveness of $\I$ modelling Artificial Neural Networks. For further uses and examples on regulation with \I, we refer the reader to \cite{garcia2010normative}

\section{$\mathcal{I^*}$: Online rule-management extension}

Continuing the work from \cite{garcia2010normative}, an extension that easily could pop up, when analysing the language and the problems one might want to solve, is the online rule-management, \emph{i.e.} the possibility of adding and removing rules on runtime. Thus, agents by means of events, and if they are accepted as valid actions, could trigger the institutional rule actions declared previously.

As for the implementation of the extension, rules  are defined as $dynamic$ Prolog clauses, as with the $\tt fired/1$ clause declared as $\tt dynamic$ after the initial comments of the code in the Appendix.

We now redefine the actions of adding and deleting open units, \emph{i.e.} constrained formulae or rules:

\begin{definition}
\label{def:s-r2}
Relation $\mathbf{s}_r(\Delta,A,\Delta')$ mapping a
state $\Delta$, the action section of a rule and a new state $\Delta'$ is defined as:
\begin{enumerate}
\item
      $\mathbf{s}_r(\Delta,(A\aconj As),\Delta')$
      holds iff both $\mathbf{s}_r(\Delta,A,\allowbreak
      \Delta_1)$ and $\mathbf{s}_r(\Delta_1, \allowbreak As,
      \allowbreak \Delta')$ hold.

\item $\mathbf{s}_r(\Delta,\add(constr\_formula),\Delta')$ holds iff 
\begin{enumerate}
	\item $constr\_formula \not\in\Delta$ and $\Delta'=\Delta \cup \{constr\_formula\}$ or;
	\item $\Delta'=\Delta$.
\end{enumerate}

\item $\mathbf{s}_r(\Delta,\add(rule),\Delta')$ holds iff 
\begin{enumerate}
	\item $rule$ does not hold and ${\tt assert}(rule)$ 
	\item $\Delta'=\Delta$.
\end{enumerate}

\item $\mathbf{s}_r(\Delta,\del(constr\_formula),\Delta')$ holds iff
\begin{enumerate}
	\item      $constr\_formula \in\Delta$ and $\Delta'=\Delta\setminus\{constr\_formula\}$ or;
	\item $\Delta'=\Delta$.
\end{enumerate}

\item $\mathbf{s}_r(\Delta,\del(rule),\Delta')$ holds iff
\begin{enumerate}
	\item      $rule$ holds and ${\tt retract}(rule)$
	\item $\Delta'=\Delta$.
\end{enumerate}	

\end{enumerate}
\end{definition}
Case 1 decomposes a conjunction and builds the new state by merging
the partial states of each update. Case 2 and 4 cater respectively for the insertion and removal of
atomic formulae $\alpha$. Finally, Case 3 and 5 formalises respectively the insertion and removal of rule $rule$, following Sictus Prolog capability to add and remove Horn clauses in runtime.

\begin{figure*}[htb]
\center

\fbox{
$
\begin{array}{rcl}
\mbox{$Rules$} & ::= & \k{rule} '(' constr\_formula,[ \mbox{$ECA$-$Rule$} \allowbreak\,|\, \mbox{$if$-$Rule$} \allowbreak\,|\,\allowbreak 
\\
&&\mbox{$ignore$-$Rule$} \allowbreak\,|\, \allowbreak
\mbox{$prevent$-$Rule$} \allowbreak\,|\,\allowbreak \mbox{$force$-$Rule$}] ')'^+
\\
\mbox{$ECA$-$Rule$} & ::= & \k{on} set\_of\_events \kb{if} conditions \kb{do} actions
\\
\mbox{$if$-$Rule$} & ::= & \k{if} conditions \kb{do} actions
\\
\mbox{$ignore$-$Rule$} & ::= & \k{ignore} set\_of\_events \kb{if} conditions 
\\
\mbox{$prevent$-$Rule$} & ::= & \k{prevent} conditions \kb{if} conditions 
\\
\mbox{$force$-$Rule$} & ::= & \k{force} set\_of\_events \kb{on} set\_of\_events\\
&& \k{if} conditions \kb{do} actions
\\
set\_of\_events & ::= & events \,|\, \emptyset
\\
events & ::= & atomic\_formula, events \,|\,
           atomic\_formula \\
conditions & ::= & conditions\conj conditions \,|\,
                     \rnot(conditions) \\
                     &\,|\, & \sat(set\_of\_constraints)      \,|\, constr\_formula	\\	
                     &\,|\, & \mathtt{seteq}(set\_of\_constraints,set\_of\_constraints)\\		 
                     &\,|\,& constraint \in set\_of\_constraints\\
                     &\,|\,& \ctime(number) \,|\, \mathtt{true} 
\\
constr\_formula & ::= & atomic\_formulae\\
			& \,|\,& atomic\_formulae : set\_of\_constraints
\\
\mathit{actions} & ::= & action\aconj actions \,|\, action
\\
open\_unit & ::= & \mbox{Rules}
\\
action & ::= & \add (open\_unit) \,|\, \del(open\_unit)
\end{array}
$ 
} \caption{Grammar for $\I^*$:Adding and Removing Rules}\label{I_star:fig:grammar} 
\end{figure*}

\section{Conclusions}\label{conclusions}

This paper poses the open question of how to regulate the computations of Artificial Neural Networks and Deep Learning and tries to open a research path towards \textbf\textit{{Artificial Teaching}} by means of  Agent-based Training for Deep Learning (ABT4DL). This proposal of Collaborative Learning use of Regulated Multi-agent Systems (MAS) for this purpose. I presented my advances along these lines, \emph{i.e.} a normative language for Regulated MAS that is used to gather a set of concurrent inputs provided by the proposed Training Agents.

The \I\ language is useful to predict a future state of affairs with an initial state and a sequence of sets of events, that occur and modify the intermediate states of affairs, until the final one is reached.
The limitations of the language are determined by the rule engine. These limitations include the inability to plan, i.e. determine the sequence of sets of events that must occur in order to reach a given state of affairs from a given initial state, or post-dicting, i.e. determine the previously unknown facts in a partial initial state given a final state and the sequence of sets of events that may have occurred previously. However, the goal of the language is to regulate a MAS and keep track of its evolution by prediction. Post-diction and planning would be interesting for a language that an agent could use for deciding which action to perform but this is not the aim of this paper. 

In this article, how agent behaviour can be regulated with the language proposed is shown, 
and it is shown that \I\ has a simple and fixed semantics that avoids normative conflicts. However different options of execution are given by adding extra ignore-rules specifying what normative positions prevail over others, in order to avoid any normative conflict when using deontic notions \cite{garcia2010normative}. 

The main contribution of $\I$ summarised from \cite{garcia2010normative} is the management of sets of events that occur simultaneously and its implementation of norms not following the standard deontic notions, \emph{i.e.}, permissions, prohibitions, and obligations. One example of application would be when a software agent wins a given good in an auction \cite{masfit}. Two options are envisaged for the payment in that scenario: \emph{1)} expect the agent to generate the event of the payment and sanction the agent if the event is not generated before a given deadline; or \emph{2)} if the Electronic Institution has control on the agent's balance, automatically generate an event of payment as if the agent would have generated it. 
In fact, an obligation (to perform an event) that may be violated is represented as the expectation of the attempts to perform it. 
However, the enforcement of an obligation (to perform a set of events) that may not be violated  is carried out by the middleware by taking these events as having been performed even they have not, denoting them as forced events. 

\section{Future Work}\label{future-work}

\subsection{Improving Computational Regulations}
As mentioned previously, the language has been tested in the Electronic Institutions (\eis) Middleware presented in \cite{estevaphd} that runs over JADE \cite{jade}. Nonetheless, the author expects to add the language to a newly developed Middleware, as it is capable of give Operational Semantics, \emph{i.e} run, a Declarative version of \ei\ protocols as it was shown in \cite{garcia2010normative}. The main advantage of a Declarative version is that it might be provided to Software Agents to reason about it using, \emph{e.g}, a BDI architecture as the one introduced in \cite{casali05}. 

\subsection{Full-Hybrid Artificial Intelligence}
However, using $\I$ as an Programmable Event-based Middle-ware opens new paths of research as it uses Hybrid AI (\i.e. mixes Autonomous Agents and Multi-agents Systems, Machine Learning, and Symbolic Programming). Furthermore, one of the main applications of Hybrid AI is in its own a whole new AI subfield, namely, Artificial Teaching. 

\subsection{Artificial Teaching}

The whole concept of Artificial Teaching is recent, and not properly defined yet. There are some mentions in the literature that I will not cite in order to engage the reader to improve the previous lines and the proposed concepts luckily exposing her results on subsequent articles.

In my humble opinion, in the research path towards General AI there are several Problem-specific milestones to reach in every sub-field of AI; and mimicking Human Intelligence and Evolution, it may seem a natural step forward to add the teaching capability to artificial learners to decrease complexity. 

Please imagine a researcher (agent) being in a continuous "Deep... and deep... and deep... and very deep... Learning" process since the beginning of its existence. To the best of my knowledge, there are very few (human) researchers (honestly, almost none) that self-learned everything on his own, with no interaction with others who may have taught him something, even involuntarily, and this happens almost every day as a Spanish proverb well says. 

\subsection{Goal Alignment}

The research question that this section tries to pose is ``How to Cooperate towards Agreements in a flawed Coordination Mechanism?". A temptative proposal that might try to answer this question would be by proposing a Left-wing and Right-wing parties Cooperation in a Pure Center-wing Goverment. From a Game theoretic point of view is a Prisoner's Dilemma on Policy-making. Current Democracy is Opinion-based, not arriving to Pure Center Agreements. The problem with current Institutions is that are very polarised in the whole spectra of Opinions and Preferences. Thus, basing Democracy in Facts on Outcomes of Policy-making actions and Learning from the past we should obtain a better Democracy where Equity and Inclusion leads the way towards maximising Social Welfare.

The Fact is there are two Nash Equilibria in the Policy-making Scenario, Left-wing and Right-wing. Then, what is a better option? Drive on the Left or drive on the Right? Thus, in one lane rural paths with no traffic signs, and where two driving directions may occur, who has traffic precedence? Current Democracy is an Opinion Fight without Cooperation. In some cases, the driving problem is easily solved by Precedence Emergence or a Car Crash may occur. Are we all willing to a Society Crash like past Economic Crashes but in a greater extent? Thus, we need to align Goals to drive in the same direction in spite of Heterogeneous Preferences of Self-interested Agents.

\subsubsection{Normative Agreements}
Our proposal is a progression for Utilitarianism\cite{sep} towards Social Utilitarianism as a Social Function Fact-based (Rules on acts)-Utilitarianism; and that, Equitable Institutions are (Social and Inclusive) Agreements, in Nash Equilibrium, on Rules of Act. Absolute Majority discriminates between winners and losers of elections, Less-Discriminatory Majority shoud be the future Goal. Current Democracy is Opinion-based, (Rules on acts)-Utilitarism should be based on Facts, i.e. Fact-Utilitarianism, Rule-Utilitarianism, Act-Utilitarianism and Preference-Utilitarianism and their inter-relationships to have a better understanding of Social Contract Theory, e.g. in which cases an institution is better than other and why; or what is a Helpful Vote. Arrow's Impossibility Theorem assumes no kind of Utility is used, so the fore-mentioned approach would avoid it, especially for Social Choice Theory. So there a need for Social Utilitarianism, i.e. Social Choice and Social Mechanism Design Theory in tandem that would avoid the limitations of Game Theory, thus starting the new field of Social Game Theory based on (Rules on acts)-Utilitarianism, i.e. a new (Social Contract and Institutions) Theory.

\subsection{Collaborative Knowledge Evolution}

With this, I want to emphasize the role of (Human and Artificial; Physical and Software) Teachers in Collaborative Learning and (Collaborative) Research, and thus in Collaborative Knowledge Evolution. Luckily, in a future we would be a step closer to General AI, by means of  Collaborative Optimisation, achieving thus a full integration and consensus of researchers (and their contributions, either Physical or Software), even they are not collaborating on purpose. And all these thanks to Regulated Middle-wares and Artificial Mediators.  

\bibliography{thesis}
\bibliographystyle{ios1.bst}

%
\end{document}